\documentclass[letterpaper]{article} % DO NOT CHANGE THIS
\usepackage{aaai25}  % DO NOT CHANGE THIS
\usepackage{times}  % DO NOT CHANGE THIS
\usepackage{helvet}  % DO NOT CHANGE THIS
\usepackage{courier}  % DO NOT CHANGE THIS
\usepackage[hyphens]{url}  % DO NOT CHANGE THIS
\usepackage{graphicx} % DO NOT CHANGE THIS
\usepackage{natbib}  % DO NOT CHANGE THIS AND DO NOT ADD ANY OPTIONS TO IT
\usepackage{caption} % DO NOT CHANGE THIS AND DO NOT ADD ANY OPTIONS TO IT
\usepackage{algorithm}
\usepackage{algorithmic}

\usepackage{multirow}
\usepackage{graphicx}
\usepackage{amsfonts,amssymb}
\usepackage[table,xcdraw]{xcolor}

\usepackage{newfloat}
\usepackage{listings}
\DeclareCaptionStyle{ruled}{labelfont=normalfont,labelsep=colon,strut=off} % DO NOT CHANGE THIS
\floatstyle{ruled}
\newfloat{listing}{tb}{lst}{}
\floatname{listing}{Listing}
\title{SalM$^2$: An Extremely Lightweight Saliency Mamba Model for Real-Time Cognitive Awareness of Driver Attention}
\author{
    Chunyu Zhao, 
    Wentao Mu, 
    Xian Zhou, 
    Wenbo Liu, 
    Fei Yan, 
    Tao Deng\thanks{Corresponding author}
}
\affiliations{
    School of Information Science and Technology, Southwest Jiaotong University, Chengdu, China%, School of Information Science and Technology‌‌
    \{zhaochunyu, mwt2858781490, zhouxian, liuwenbo\}@my.swjtu.edu.cn, 
    \\fyan@home.swjtu.edu.cn, tdeng@swjtu.edu.cn
}

\usepackage{bibentry}
\begin{document}

\maketitle

\begin{abstract}
Driver attention recognition in driving scenarios is a popular direction in traffic scene perception technology. It aims to understand human driver attention to focus on specific targets/objects in the driving scene. However, traffic scenes contain not only a large amount of visual information but also semantic information related to driving tasks. Existing methods lack attention to the actual semantic information present in driving scenes. Additionally, the traffic scene is a complex and dynamic process that requires constant attention to objects related to the current driving task. Existing models, influenced by their foundational frameworks, tend to have large parameter counts and complex structures. Therefore, this paper proposes a real-time saliency Mamba network based on the latest Mamba framework. As shown in Figure \ref{fig:highlight_compare}, our model uses very few parameters (0.08M, only 0.09$\sim$11.16\% of other models), while maintaining SOTA performance or achieving over 98\% of the SOTA model's performance.
% Our code is available at the github link https://github.com/zhao-chunyu/SaliencyMamba. 
% \link{Code}{https://github.com/zhao-chunyu/SaliencyMamba}
% Therefore, based on the latest network framework Mamba, this paper proposes a real-time saliency Mamba network that uses very few parameters (0.08M, only 0.09\% to 11.16\% of other models) while maintaining state-of-the-art (SOTA) performance, or achieving \>98\% of the performance of SOTA models.
\end{abstract}

\begin{links}
    \link{Code}{https://github.com/zhao-chunyu/SaliencyMamba}
\end{links}

\section{Introduction}
With the rapid advancement of autonomous driving technology, understanding and predicting driver behavior has become increasingly important. Among the many factors affecting driving safety, the driver's attention state is crucial \cite{balasubramani2024ddss}. Distraction or insufficient attention can lead to reduced perception of the surrounding environment, increasing the risk of traffic accidents \cite{fang2024behavioral}. Therefore, research on methods for recognizing driver attention is of great significance for enhancing road safety and advancing the development of intelligent driving assistance systems. 
\begin{figure}[htp]
    \centering
    \includegraphics[width=1\linewidth]{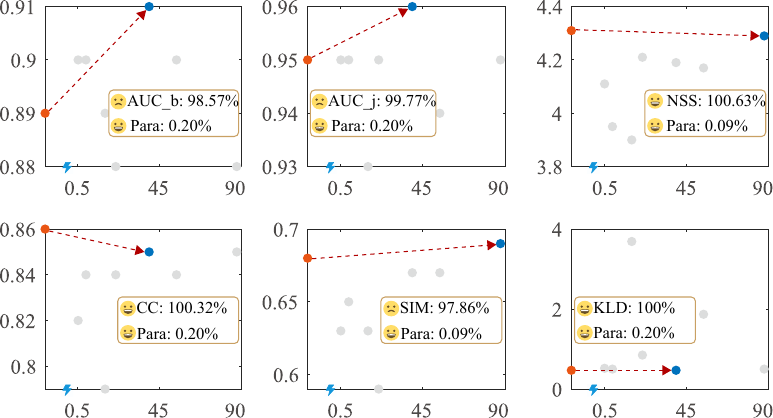}
    \caption{Comparison of Parameters and Performance. We compare our model with other state-of-the-art models on the DrFixD-rainy dataset. The horizontal axis represents the number of parameters (M), and the vertical axis represents performance. \textcolor[rgb]{0.9176, 0.3333, 0.0784}{$\bullet$} denotes our model, and \textcolor[rgb]{0, 0.4471, 0.7412}{$\bullet$} denotes the best-performing model among the comparison models, and \textcolor[rgb]{0.8627, 0.8667, 0.8667}{$\bullet$} denotes other comparison models. \includegraphics[width=0.012\textwidth]{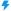} denotes truncated coordinates, with the origin set to 0.08.}
    \label{fig:highlight_compare}
\end{figure}

Given that the traffic driving scenario is dynamic and complex \cite{alkaabi2023identification}, it is unwise to analyze driver attention solely based on features from the driving scene. The driving environment contains not only rich ``Bottom-up" image features (such as the sky, buildings, vehicles, etc.) but also ``Top-down" information related to the current driving task (such as vehicles ahead, pedestrians crossing the road, traffic lights, etc.). Relying only on features within the driving scene to identify driver attention may cause the model to overfit to the labeled driver attention data, leading to a lack of understanding of the driving context and reducing the model's generalization ability. However, within the visual attention system, relevant stimuli and cues can be selected while filtering out less relevant ones from the rich visual information around us. Attention resources are then allocated to these stimuli and cues, typically by directing gaze toward objects of interest in the visual environment \cite{hertz2023aberrant,itti2001computational}. This process is known as selective visual attention or selective attention allocation \cite{evans2011visual}. Similarly, in specific traffic driving scenarios, a driver's visual attention, driven by the driving task, tends to focus on one or more regions or targets that are related to the task at hand. When performing a driving task, a driver’s attention needs to remain highly focused, consistently allocating attention to objects or areas relevant to the current driving task. If the driving scenario or task changes, the driver’s attention will also shift accordingly. Therefore, the semantic information of the current driving scene is crucial for understanding the subsequent shifts in driver attention.

Numerous researchers have studied driver attention prediction, but most of these studies focus on the overall features of the driving scene, often neglecting the understanding of semantic information present within the scene \cite{xia2019predicting,baee2021medirl,shen2022cocatt}. Additionally, existing models tend to have large parameter counts and high computational complexity \cite{fang2022dada,chen2023fblnet}. Therefore, this paper proposes a novel driver attention recognition model based on driving events, which not only has a small parameter count but also utilizes the semantic information from the driving scene to guide the recognition of current driver attention. Specifically, we employ a dual-branch structure: one ``Bottom-up" branch extracts image features from the driving scene, while the other ``Top-down" branch captures the semantic information of the scene. At the deepest layer of image feature extraction, we integrate the semantic information of the driving scene to guide the image features. This semantic-guided feature is then passed through a decoder module to obtain the final driver attention map. Since our driver gaze prediction is driven by the contextual information of the driving scene, the resulting attention maps align more closely with the actual distribution of driver attention. 
% Through comprehensive research into understanding driving scenes and recognizing driver attention, we aim to provide more precise and real-time driver attention monitoring tools for intelligent transportation systems, ultimately contributing to a safer and smarter driving environment. 
To summarize, the contributions of this work are the following:
\begin{itemize}
\item We propose a driver attention prediction architecture inspired by human visual attention mechanism. Our architecture include a ``Bottom-up" branch that extracts stimulus-driven image features from the driving scene, and a ``Top-down" branch that extracts task-driven semantic features. The ``Top-down" branch includes a novel Selective Channel Parallel Mamba (SCPM) layer, which not only addresses the parameter explosion issues caused by high-dimensional data in Mamba, but also corrects the information loss from fixed channel partition.
\item We design a novel Cross-Modal Attention (CMA) fusion module. This module integrates the semantic information from the cognitive process and the image features to guide the allocation of driver attention in driving scenarios. Our module adopts the idea of channel attention to address the challenge of feature dimension mismatch between semantic and image information, allowing for effective fusion. The entire module introduces only a single additional parameter.
\item We develop an extremely lightweight effective driver attention prediction network. To the best of our knowledge, the SalM$^2$ network is the most lightweight model for driver saliency prediction, with only 0.08M trainable parameters. The model is trained on three popular datasets, and SalM$^2$ achieves SOTA performance using only a fraction (approximately 0.09\% to 11.16\%) of the parameters compared to other models, or reaches 98+\% of the performance of the SOTA models.
\end{itemize}

\section{Related Work}
\subsection{Driver Saliency Prediction}
% In the field of driver saliency prediction, Tawari et al. improved gaze estimation by collecting first-person driving perspectives using Google Glass \cite{tawari2014attention}. Deng et al. proposed a bottom-up saliency detection model that predicts driver gaze points through both low-level and high-level features  \cite{deng2017learning}. Traditional models often focus on static driving scenes and lack dynamic understanding of real driving conditions. Alletto et al. released the DR(eye)VE dataset  \cite{alletto2016dr}, which includes a variety of weather and driving conditions, enhancing the understanding of driver attention processes. Xia et al. constructed the BDD-A dataset \cite{xia2019predicting}, which includes emergency braking events, and introduced the Human Weighted Sampling (HWS) method for predicting driver attention. Fang et al. published the DADA-2000 dataset \cite{fang2019dada}, covering normal driving and accident scenarios, and proposed a multi-path semantic-guided attention fusion network. Deng et al. released the TrafficGaze dataset  \cite{deng2019drivers}, providing richer driving scene data. Tian and Deng et al. developed the DrFixD-rainy/night dataset \cite{tian2022driving,deng2023driving}, addressing the research gap in complex weather conditions. These datasets and methods continue to drive advancements in the field of driver gaze prediction.
Previous works have made notable advances in driver saliency prediction, primarily focusing on inherent visual features or semantic information from image segmentation and optical flow. Tawari et al. pioneered first-person gaze estimation using Google Glass \cite{tawari2014attention}, while Deng et al. proposed a bottom-up saliency model combining low and high-level features \cite{deng2017learning}, though limited by traditional machine learning's feature extraction capabilities.

As research progressed, numerous datasets and deep learning algorithms emerged to address conventional limitations. The DR(eye)VE dataset by Alletto et al. covers various driving conditions but lacks scene diversity and semantic richness \cite{alletto2016dr}. The BDDA dataset introduced by Xia et al. enriches the field with urban driving scenarios and emergency events \cite{xia2019predicting}. Fang et al. presented DADA-2000 dataset covering normal and accident scenarios with a semantic-guided attention fusion network, though limited to collision scenes and segmentation-based semantics \cite{fang2019dada}. The TrafficGaze dataset by Deng et al. offers comprehensive clear-weather data with a lightweight CNN framework \cite{deng2019drivers}, while the DrFixD-rainy/night dataset specifically addresses adverse weather conditions \cite{tian2022driving,deng2023driving}. Nevertheless, these approaches rely on scene features or segmentation-based semantics, lacking scene understanding. Brishtel et al. demonstrated correlations in gaze patterns across driving modes \cite{9945234}. Vozniak et al. successfully incorporated semantic danger cues in attention prediction \cite{vozniak2023context}, though obtaining annotated semantic data remains challenging.

In summary, while substantial research has developed numerous models for predicting driver attention, there remains a gap in utilizing basic semantic guidance of driving scenes for driver saliency prediction. Therefore, this paper employs the CLIP \cite{radford2021learningtransferablevisualmodels} model to extract semantic information from driving scenes. To validate the effectiveness of our proposed method in diverse and dynamic driving scenarios, we conduct experiments across datasets with different weather conditions (TrafficGaze, DrFixD-rainy) and complex semantic information (BDDA).

\subsection{Downstream Tasks}
\subsubsection{Salient Object Detection.} This is an important task in computer vision aimed at detecting the most prominent object regions in an image. In the context of traffic driving scenes, drivers often automatically filter out objects unrelated to the current driving task. Therefore, some existing works use driver attention allocation as prior knowledge to detect prominent or key objects \cite{qin2022id,shi2023fixated}. This approach does not detect all objects in the driving scene but focuses on those most relevant to the current driving task, thereby reducing redundant information.
\subsubsection{Drive Event Recognition.} In the field of intelligent transportation systems, detecting driving events by analyzing driving scene information is one of the key tasks in preventing traffic accidents. Driver visual attention helps identify information relevant to the current driving task while suppressing irrelevant information. Du et al. \cite{du2023causes} established an attention-based driving event dataset (ADED) and proposed a driver attention-guided model that uses driver attention as guidance to better recognize events that cause shifts in driver attention.

In conclusion, we believe that the semantic information of the driving scene is beneficial in identifying the driver’s attention most aligned with the current driving task. Inspired by visual cognition, we have developed an attention prediction model driven by the semantic information of the current driving scene, integrating this semantic information at the deepest level of image feature extraction to guide the driver's attention effectively.
\begin{figure*}
    \centering
    \includegraphics[width=1\linewidth]{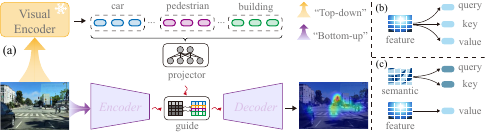}
    \caption{Overview of the proposed SalM$^2$ network. (a) shows the overall network framework, which includes two branches: a ``Bottom-up" branch and a ``Top-down" branch. (b) illustrates the principle of the self-attention mechanism. (c) illustrates the principle of our proposed cross-modal attention mechanism.}
    \label{fig:model_overview}
\end{figure*}
\section{Proposed Method}

Since the traffic scene is a complex and dynamically changing setup, the driver requires continuous focus on objects/areas that are significantly relevant to the current driving scene. However, most of the previous work only considered spatio-temporal information in driving scenarios. In fact, solely relying on visual information from the driving scene is not sufficient to accurately identify the locations or areas that need the driver's attention \cite{rasouli2018joint} — that is the ``Bottom-up" image information. This is because the driver's attention is closely related to the current driving task, and it is also necessary to consider the ``Top-down" semantic information related to the current driving task that is embedded in the driving scene.

% \subsection{SalM$^2$: Saliency model based mamba}
\subsubsection{Overall Network Framework.}
Considering these elements, we propose a saliency mamba model, named SalM$^2$ that uses ``Top-down" driving scene semantic information to guide ``Bottom-up" driving scene image information to simulate human drivers' attention allocation. The overall framework is illustrated in Figure \ref{fig:model_overview}, our model first utilizes an efficient network framework Mamba structure to build an extremely lightweight backbone network for extracting ``Bottom-up" features from image and decoding the features. Simultaneously, a visual encoder is used to understand the ``Top-down" semantic information driven by the driving task in the current traffic scene. Then, when the deepest features representing image information of the driving scene is extracted from the backbone network, we utilize the ``Top-down" semantic information to guide the ``Bottom-up" image information. Finally, we decode the perceived information in the decoder of the backbone network to obtain the driver attention map.

% \subsection{SalM: Efficient attention network}
\subsubsection{Reliable Attention Prediction Network.}
\begin{figure} 
    \centering
    \includegraphics[width=1\linewidth]{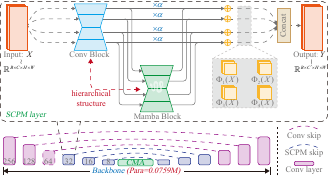}
    \caption{Illustrate of the ``Bottom-up" backbone network. The purple and blue dashed lines represent skip connections, which include both spatial and channel attention. The skip connections share weights.}
    \label{fig:backbone}
\end{figure}

Since we aim for intelligent vehicles to have the capability to focus on objects related to driving tasks in the scene, just like a driver, we need to develop a network with low computational cost and deployability with lower hardware requirements. To achieve this goal, we construct an extremely lightweight backbone network as the image encoding and decoding network, as shown in Figure \ref{fig:backbone}. This design allows us to fully utilize the Mamba structure's efficient feature representation capability, accelerating the learning of the representation space at the channel level. Inspired by the work of Wu et al. \cite{wu2024ultralight}, we modified it to perform high-dimensional feature representation in our selective channel parallel Mamba (SCPM) layers. This framework contains only 0.0759M parameters, with a model size of just 1.6MB. 

In this architecture, the overall backbone network is a hierarchical structure based on Mamba \cite{gu2023mamba}. We first use convolutional network layers to extract low-level features from images, and then employ SCPM layers to further extract high-level features. To leverage the efficiency of the Mamba framework and address the parameter disaster caused by high-dimensional data, a typical solution is to split the input $X\in\mathbb{R}^{B{\times}C{\times}H{\times}W}$ into four parts ${x_i}\in\mathbb{R}^{B{\times}C/4{\times}H{\times}W},i=1,2,3,4$ and then process them in parallel through the Mamba layers. To correct the information loss caused by fixed channel splitting, we design the SCPM layer. This layer embeds the input features into convolutional blocks in parallel and then feeds them into the Mamba layer in parallel. The output feature channels of the parallel branches differ, but the feature dimensions remain consistent. The operation of each parallel channel is as shown in Equation \ref{eq:scpm}.
\begin{equation}
\Phi_{i}(X)=Con\nu_{i}(X)+{\alpha\times}Mamba(Con\nu_{i}(X)),
\label{eq:scpm}
\end{equation}
\begin{equation}
Y=\sum_{i=1}^{i=n}\Phi_{i}(X),i\in\{1,2,3,4\},n=4,
\end{equation}
where $\Phi_{i}(X)$ represents the output of the $i^{th}$ parallel layer, $\alpha$ is a skip scale, $Y\in\mathbb{R}^{B{\times}C'{\times}H{\times}W}$ represents the final output of the SCPM layer, and $\sum$ represents the summation operation, which in this context is the summation along the channel dimension.

\subsubsection{Cross-Modal Attention Mechanism.}
For most models, the backbone network is used to extract ``Bottom-up'' image information to predict driver attention \cite{min2019tased,9428151,tian2022driving,chen2023fblnet}. However, the traffic environment is a complex and dynamic process where different driving scenarios involve specific driving tasks (such as stopping at traffic lights, yielding to pedestrians crossing the road, etc.). Therefore, we need to understand the specific driving tasks in the current driving scenario. Using this ``Top-down" scene information to guide the ``Bottom-up" image information can better identify the driver's attention allocation. However, text data contains strong semantic and logical properties, making it difficult to match and align with image data in the feature space. To further explore how to use semantic information to guide driving scene information, our proposes using a cross-modal attention mechanism (CMA).
% Existing large language models already possess the capability to understand images, incorporating rich semantic information when analyzing them. We utilize the CLIP model for extracting semantic information. We then apply our proposed cross-modal attention mechanism to fuse the information from both modalities, enabling the semantic information to guide the image information. The specific implementation process of the attention mechanism is illustrated in Figure 3.

\begin{figure}
    \centering
    \includegraphics[width=1\linewidth]{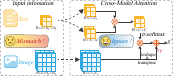}
    \caption{Illustrate of the cross-modal attention mechanism. In the figure, $\otimes$ denotes matrix multiplication, and $\oplus$ represents element-wise addition.}
    \label{fig:CMA}
\end{figure}

As shown in Figure \ref{fig:CMA}, CMA uses the semantic information $Info_{text}$ extracted by CLIP \cite{radford2021learningtransferablevisualmodels} and the image information $Info_{image}$ extracted by the backbone network as inputs. However, directly fusing these inputs leads to a dimension mismatch problem as $\mathbb{D}_{im}(Info_{text})\neq\mathbb{D}_{im}(Info_{image})$. Therefore, we project the semantic information into the same feature channel space as the image information. Using the simple structure of the channel attention module \cite{fu2019dual}, we achieve cross-modal information fusion, thereby avoiding the issue of scale mismatch between different types of information. 

Due to CLIP being a large model trained through contrastive learning, the extracted semantic information and image features have already been aligned in a similar feature space through contrastive training, enabling us to readily obtain the original semantic information $Info_{text}^{ori}\in\mathbb{R}^{\mathcal{B}{\times}{Token}}$ from images. Nevertheless, we still need to perform a easy dimensional projection on the semantic information to ensure identical channel dimensions between the two modalities while preserving all information. we map the original semantic information $Info_{text}^{ori}$ to the image feature space, obtaining new semantic information $Info_{text}$.
% \in\mathbb{R}^{\mathcal{B}{\times}\mathcal{C}{\times}\mathcal{H}_1{\times}\mathcal{W}_1}
\begin{equation}
Info_{text}=projector(Info_{text}^{ori}).
\end{equation}

Within the CMA module: we first reshape the semantic information $Info_{text}\in\mathbb{R}^{\mathcal{B}{\times}\mathcal{C}{\times}\mathcal{H}_1{\times}\mathcal{W}_1}$ to $Q\in\mathbb{R}^{\mathcal{B}{\times}\mathcal{C}{\times}\mathcal{N}_1}$ as Query and transpose $Info_{image}$ to $K\in\mathbb{R}^{\mathcal{B}{\times}\mathcal{N}_1{\times}\mathcal{C}}$ as Key. Then, we perform a matrix multiplication between the $Q$ and the $K$. Finally, we apply a softmax operation to obtain the saliency semantic representation $S\in\mathbb{R}^{\mathcal{B}{\times}\mathcal{C}{\times}\mathcal{C}}$.
\begin{equation}
s_{ji}=\frac{exp(Q_i,K_j)}{\sum_{i=1}^\mathcal{C}exp(Q_i,K_j)},
\end{equation}
where $s_{ji}$ measures the influence of the $i^{th}$ channel on the $j^{th}$ channel, $\mathcal{C}$ is the total number of channels, and $exp(\cdot,\cdot)$ is used to measure the similarity between two channels.

In addition, we reshape the deepest image information $Info_{image}\in\mathbb{R}^{\mathcal{B}{\times}\mathcal{C}{\times}\mathcal{H}_2{\times}\mathcal{W}_2}$ to $V\in\mathbb{R}^{\mathcal{B}{\times}\mathcal{C}{\times}\mathcal{N}_2}$ as the value and perform a matrix multiplication between $S$ and $V$, then reshape the result to $Info_{fusion}\in\mathbb{R}^{\mathcal{B}{\times}\mathcal{C}{\times}\mathcal{H}_2{\times}\mathcal{W}_2}$. Next, we multiply the attention $S$ by a scaling parameter $\gamma$ and add the image feature information $Info_{image}$ to obtain the final cross-modal fused information $S'\in\mathbb{R}^{\mathcal{B}{\times}\mathcal{C}{\times}\mathcal{H}_2{\times}\mathcal{W}_2}$.

\begin{equation}
S'_j=\gamma\sum_{i=1}^\mathcal{C}(s_{ji}\cdot{V_i})+Info_{image}^j,
\end{equation}
where $\gamma$ is a learnable weight and is initialized to 0, $\mathcal{N}_1=\mathcal{H}_1\times\mathcal{W}_1$, $\mathcal{N}_2=\mathcal{H}_2\times\mathcal{W}_2$, $\mathcal{H}_2=2\mathcal{H}_1$, and $\mathcal{W}_2=2\mathcal{W}_1$.
\begin{figure*}
    \centering
    \includegraphics[width=1\linewidth]{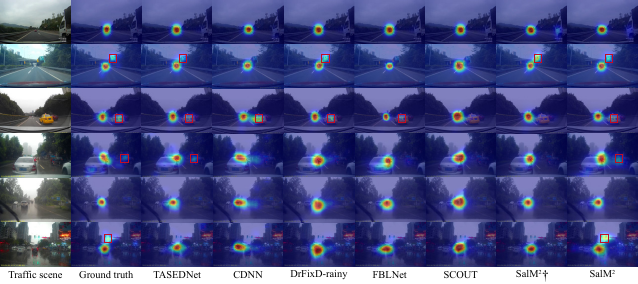}
    \caption{Qualitative evaluation comparison of our model and the other SOTA methods. Since the weight files of MLNet and SCAFNet are not available in original papers, we could not perform visualization comparisons under identical conditions. {$\dagger$ is SalM$^2$ without branch of semantic.} \textcolor[rgb]{1, 0, 0}{$\square$} delineates the attention allocation regions influenced by semantic information.} 
    \label{fig:compare1}
\end{figure*}
\section{Experiment}
\subsubsection{Datasets.}
In this work, we train our proposed SalM$^2$ model on 3 representative driver attention datasets: TrafficGaze, DrFixD-rainy, and BDD-A. Before training, we resize the traffic images in each dataset to 256$\times$256 to reduce the computational load on the network. Additionally, the compared models are divided into static driver attention prediction methods and dynamic driver attention prediction methods. In the static methods, the model relies on a single-frame driving scene image to predict driver attention. In the dynamic driver attention methods, we maintain the original method's sequence length as the input.

\subsubsection{Implementation Details.}
The ``Top-down" semantic information in this work is initialized and extracted using CLIP, with the pre-trained model being RN50$\times$16. The model parameters are entirely frozen and do not impact the actual training process, and thus are not included in the actual parameter count. We choose the Adam optimizer with a learning rate set to $10^{-3}$, momentum of 0.9, and weight decay of $10^{-4}$. The model training is conducted on a GPU server equipped with an NVIDIA GeForce RTX 4090 24GB GPU and an Intel(R) Xeon(R) Silver 4314 CPU @ 2.40GHz. It is worth noting that for the SCOUT network \cite{kotseruba2024scout+} used in the comparison, we used the version without additional annotations as described in the original paper, since the original SCOUT network involved extra dataset annotations. 
% Our loss function employs the Binary Cross-Entropy Loss (BCELoss).

\subsubsection{Loss function.}
The loss function $\mathcal{L}_{sal}$ is defined for predicting the saliency map $\hat{S}$ and the ground truth saliency map $S$. The $\mathcal{L}_{sal}$ is the BCE loss, which is formulated as follows.
\begin{equation}
    \mathcal{L}_{sal}(S,\hat{S})=-\frac{1}{N}\sum_{i=1}^{N}S_{i}\log(\hat{S})+(1-S_{i})\log(1-\hat{S}_{i})
\end{equation}

\section{Results}
To evaluate the performance of our proposed model, we conduct both qualitative and quantitative comparisons with 7 other popular methods. These include 4 static methods: MLNet \cite{cornia2016deep}, CDNN \cite{deng2019drivers}, FBLNet \cite{chen2023fblnet}  and SCOUT \cite{kotseruba2024scout+}) and 3 dynamic methods: TASED-Net \cite{min2019tased}, DrFixD-rainy \cite{tian2022driving} and SCAFNet \cite{fang2021dada}.
\subsection{Qualitative Evaluation} 
Figure \ref{fig:compare1} presents a qualitative comparison with the SOTA models. For better visualization, we overlay the predicted saliency maps on the original images. Overall, we can conclude that the SalM$^2$ model demonstrates superior predictive performance compared to existing saliency models. In rows 1, and 5 of Figure \ref{fig:compare1}, we can see that existing saliency models exhibit good prediction performance for traffic lights, cars, and some road lanes in traffic scenes. However, these models fail to capture the most crucial ``Top-down" information related to driving, and their results do not align well with standard saliency map. For example, in rows 2, 3, 4, and 6 of Figure \ref{fig:compare1}, the ``Top-down" information related to the current driving task (such as road signs, nearby vehicles, and traffic lights) is better simulated by our model compared to other models in terms of capturing human driver attention.

\subsection{Quantitative Evaluation}
We use six metrics to evaluate the performance of different prediction methods. These include five similarity metrics: Area Under the ROC Curve (AUC\_Borji \cite{judd2009learning} and AUC\_Judd \cite{borji2012quantitative}), Normalized Scanpath Saliency (NSS \cite{peters2005components}), Linear Correlation Coefficient (CC \cite{le2007predicting}), Similarity Metric (SIM \cite{borji2012quantitative})  and one dissimilarity metric: Kullback-Leibler Divergence (KLD \cite{riche2013saliency}). 

\begin{table} 
\centering
\fontsize{10}{12}\selectfont%设置字体 {大小}{间距}
\resizebox{\linewidth}{!}{%
\begin{tabular}{c|ccc|cc}
\hline % 0.14 0.12 2.68
Model        & CC↑           & SIM↑          & KLD↓          & Params (M) & FLOPs (G) \\ \hline\hline
MLNet        & 0.61          & 0.43          & 1.20          & 15.45      & 34.46     \\
TASED-Net    & 0.55          & 0.42          & 1.24          & 21.26      & 91.48     \\
CDNN         & 0.62          & 0.45          & 1.14          & \underline{0.68}       & \underline{7.10}      \\
SCAFNet      & \textbf{0.64} & \textbf{0.48} & 1.09          & 54.51      & 85.86     \\
DrFixD-rainy & \textbf{0.64} & \underline{0.47}          & 1.09          & 39.57      & 9.74      \\
FBLNet       & \textbf{0.64} & \underline{0.47}          & 1.40          & 87.48      & 20.31     \\
SCOUT        & \underline{0.63}          & \textbf{0.48} & \textbf{1.04} & 4.96       & 17.72     \\ \hline\hline
SalM$^2$ &
  \cellcolor[HTML]{EFEFEF}\textbf{0.64} &
  \cellcolor[HTML]{EFEFEF}\underline{0.47} &
  \cellcolor[HTML]{EFEFEF}\underline{1.08} &
  \cellcolor[HTML]{EFEFEF}\textbf{0.08} &
  \cellcolor[HTML]{EFEFEF}\textbf{4.45} \\ \hline
 % &
 %  \cellcolor[HTML]{EFEFEF}0.00 &
 %  \cellcolor[HTML]{EFEFEF}-0.01 &
 %  \cellcolor[HTML]{EFEFEF}-0.04 &
 %  \cellcolor[HTML]{EFEFEF}0.60 &
 %  \cellcolor[HTML]{EFEFEF}2.65 \\ \hline
\end{tabular}%
}
\caption{Performance comparison on BDDA using saliency evaluation metrics. Symbol ↑ expects a larger value and ↓ expects a smaller value. Best scores are shown in \textbf{bold}, the second best is \underline{underlined}.}
\label{tab:compare1}
\end{table}

Table \ref{tab:compare1} presents the quantitative performance metrics obtained by training the proposed model and other saliency models on the BDD-A dataset. As shown in the Table \ref{tab:compare1}, the proposed model achieves SOTA performance or performance close to SOTA while using only approximately 11\%, or even as little as 0.1\%, of the parameters of other models. Additionally, our model exhibits significantly lower FLOPs compared to other models, resulting in lower computational costs and making it more suitable for deployment. Therefore, it can be concluded that the proposed SalM$^2$ prediction model demonstrates superior performance compared to other models.

Simultaneously, we further validated the model on two additional datasets, TrafficGaze and DrFixD-rainy, and compared it against the same algorithms. As shown in Tables \ref{tab:compare2} and \ref{tab:compare3}, our model achieved either state-of-the-art (SOTA) or second-to-SOTA performance on all but one metric within the same dataset, while using only 11\% to 0.1\% of the parameters of other SOTA models.

\begin{table} 
\centering
\fontsize{10}{12}\selectfont%设置字体 {大小}{间距}
\resizebox{\linewidth}{!}{%
\begin{tabular}{c|cccccc}
\hline
Model &
  \begin{tabular}[c]{@{}c@{}}AUC\_\\ Borji↑\end{tabular} &
  \begin{tabular}[c]{@{}c@{}}AUC\_\\ Judd↑\end{tabular} &
  NSS↑ &
  CC↑ &
  SIM↑ &
  KLD↓ \\ \hline\hline
MLNet        & 0.87 & 0.90          & 5.69          & 0.87 & 0.45          & 0.87          \\
TASED-Net    & \underline{0.92} & \underline{0.97}          & 5.73          & \underline{0.94} & \textbf{0.79} & 1.43          \\
CDNN         & \textbf{0.93} & \underline{0.97}          & 5.83          & \textbf{0.95} & \underline{0.78}          & \underline{0.29}         \\
SCAFNet      & -    & \textbf{0.98} & \underline{6.10}          & \underline{0.94} & 0.77          & 0.66          \\
DrFixD-rainy & \underline{0.92} & \textbf{0.98} & 6.01          & \underline{0.94} & \underline{0.78}          & \textbf{0.28} \\
FBLNet       & 0.87 & \underline{0.97}          & \textbf{6.50} & 0.90 & 0.69          & 0.46          \\
SCOUT        & 0.91 & \underline{0.97}          & 5.35          & 0.91 & 0.72          & 0.39          \\ \hline\hline
SalM$^2$ &
  \cellcolor[HTML]{EFEFEF}\underline{0.92} &
  \cellcolor[HTML]{EFEFEF}\textbf{0.98} &
  \cellcolor[HTML]{EFEFEF}5.90 &
  \cellcolor[HTML]{EFEFEF}\underline{0.94} &
  \cellcolor[HTML]{EFEFEF}\underline{0.78} &
  \cellcolor[HTML]{EFEFEF}\textbf{0.28} \\ \hline
 % &
 %  \cellcolor[HTML]{EFEFEF}-0.01 &
 %  \cellcolor[HTML]{EFEFEF}0.00 &
 %  \cellcolor[HTML]{EFEFEF}-0.60 &
 %  \cellcolor[HTML]{EFEFEF}-0.01 &
 %  \cellcolor[HTML]{EFEFEF}-0.01 &
 %  \cellcolor[HTML]{EFEFEF}0.00 \\ \hline
\end{tabular}%
}
\caption{Performance comparison on TrafficGaze using saliency evaluation metrics. The symbol ``-" indicates that this metric was not computed officially.}
\label{tab:compare2}
\end{table}

\begin{table}
\centering
\fontsize{10}{12}\selectfont%设置字体 {大小}{间距}
\resizebox{\linewidth}{!}{%
\begin{tabular}{c|cccccc}
\hline
Model &
  \begin{tabular}[c]{@{}c@{}}AUC\_\\ Borji↑\end{tabular} &
  \begin{tabular}[c]{@{}c@{}}AUC\_\\ Judd↑\end{tabular} &
  NSS↑ &
  CC↑ &
  SIM↑ &
  KLD↓ \\ \hline\hline
MLNet        & 0.89          & 0.93          & 3.90 & 0.79 & 0.63          & 3.69          \\
TASED-Net    & 0.88          & \underline{0.95}          & 4.21 & 0.84 & 0.59          & 0.85          \\
CDNN         & \underline{0.90}          & \underline{0.95}          & 4.11 & 0.82 & 0.63          & 0.52          \\
SCAFNet      & \underline{0.90}          & 0.94          & 4.17 & 0.84 & 0.67          & 1.87          \\
DrFixD-rainy & \textbf{0.91} & \textbf{0.96} & 4.19 & \underline{0.85} & 0.67          & \textbf{0.47} \\
FBLNet       & 0.88          & \underline{0.95}          & \underline{4.29} & \underline{0.85} & \textbf{0.69} & \underline{0.50}          \\
SCOUT        & \underline{0.90}          & \underline{0.95}          & 3.95 & 0.84 & 0.65          & \underline{0.50}          \\ \hline\hline
SalM$^2$ &
  \cellcolor[HTML]{EFEFEF}0.89 &
  \cellcolor[HTML]{EFEFEF}\underline{0.95} &
  \cellcolor[HTML]{EFEFEF}\textbf{4.31} &
  \cellcolor[HTML]{EFEFEF}\textbf{0.86} &
  \cellcolor[HTML]{EFEFEF}\underline{0.68} &
  \cellcolor[HTML]{EFEFEF}\textbf{0.47} \\ \hline
 % &
 %  \cellcolor[HTML]{EFEFEF}-0.02 &
 %  \cellcolor[HTML]{EFEFEF}-0.01 &
 %  \cellcolor[HTML]{EFEFEF}0.02 &
 %  \cellcolor[HTML]{EFEFEF}0.01 &
 %  \cellcolor[HTML]{EFEFEF}-0.01 &
 %  \cellcolor[HTML]{EFEFEF}0.00 \\ \hline
\end{tabular}%
}
\caption{Performance comparison on DrFixD-rainy using saliency evaluation metrics.}
\label{tab:compare3}
\end{table}

\subsection{Ablation Study}

\subsubsection{Reliable Attention Prediction Network.} The driving scene is a dynamic environment that requires real-time feedback, necessitating an extremely efficient network for predicting driver attention. Inspired by the work of Wu et al. on medical image segmentation, we designed a lightweight backbone network better suited for predicting driver gaze points. To ensure fairness, we compare the backbone network of SalM$^2$ without the semantic information branch against the original network on the TrafficGaze, DrFixD-rainy, and BDD-A datasets.

As shown in Table \ref{tab:study_1}, our model demonstrates leading performance across all three datasets. Except for the AUC\_Borji metric on the DrFix-rainy dataset, our model achieves superior performance on all other metrics. Therefore, our backbone network represents a more efficient backbone network for predicting driver gaze points.
\begin{table} 
\centering
\fontsize{14}{18}\selectfont%设置字体 {大小}{间距}
\resizebox{\linewidth}{!}{%
\begin{tabular}{c|c|cccccc}
\hline
Dataset & Model & \begin{tabular}[c]{@{}c@{}}AUC\_\\ Borji↑\end{tabular} & \begin{tabular}[c]{@{}c@{}}AUC\_\\ Judd↑\end{tabular} & NSS↑ & CC↑  & SIM↑ & KLD↓ \\ \hline\hline
        & Wu et al.   & \textbf{0.93}                                          & \textbf{0.98}                                         & 5.82 & 0.93 & 0.77 & 0.31 \\
\multirow{-2}{*}{TraffiGaze} &
  \cellcolor[HTML]{EFEFEF}Backbone &
  \cellcolor[HTML]{EFEFEF}\textbf{0.93} &
  \cellcolor[HTML]{EFEFEF}\textbf{0.98} &
  \cellcolor[HTML]{EFEFEF}\textbf{5.87} &
  \cellcolor[HTML]{EFEFEF}\textbf{0.94} &
  \cellcolor[HTML]{EFEFEF}\textbf{0.78} &
  \cellcolor[HTML]{EFEFEF}\textbf{0.28} \\ \hline
        & Wu et al.    & \textbf{0.90}                                          & \textbf{0.95}                                         & 4.21 & 0.85 & 0.67 & 0.47 \\
\multirow{-2}{*}{\begin{tabular}[c]{@{}c@{}}DrFixD\\ -rainy\end{tabular}} &
  \cellcolor[HTML]{EFEFEF}Backbone &
  \cellcolor[HTML]{EFEFEF}0.89 &
  \cellcolor[HTML]{EFEFEF}\textbf{0.95} &
  \cellcolor[HTML]{EFEFEF}\textbf{4.29} &
  \cellcolor[HTML]{EFEFEF}\textbf{0.87} &
  \cellcolor[HTML]{EFEFEF}\textbf{0.69} &
  \cellcolor[HTML]{EFEFEF}\textbf{0.46} \\ \hline
        & Wu et al.    & -                                                      & -                                                     & -    & 0.62 & 0.45 & 1.12 \\
\multirow{-2}{*}{BDDA} &
  \cellcolor[HTML]{EFEFEF}Backbone &
  \cellcolor[HTML]{EFEFEF}- &
  \cellcolor[HTML]{EFEFEF}- &
  \cellcolor[HTML]{EFEFEF}- &
  \cellcolor[HTML]{EFEFEF}\textbf{0.63} &
  \cellcolor[HTML]{EFEFEF}\textbf{0.47} &
  \cellcolor[HTML]{EFEFEF}\textbf{1.09} \\ \hline
\end{tabular}%
}
\caption{Ablation study of the backbone network. The symbol ``-" indicates that this metric could not be computed due to reasons related to the dataset.}
\label{tab:study_1}
\end{table}

\begin{figure}
    \centering
    \includegraphics[width=1\linewidth]{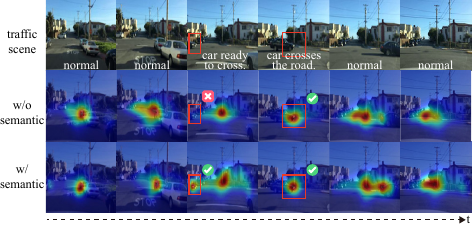}
    \caption{Visualize the driver's attention under the semantic information of a car crossing the road. White text represents semantic information, with ``normal" indicating that the driver is looking ahead and no semantic information.}
    \label{fig:example1}
\end{figure}
\subsubsection{Effective Semantic Information.} In complex driving scene, the concentration of driver attention is a key factor in ensuring road safety. By understanding how specific tasks guide attention, we can design more effective driver assistance systems to improve road safety. In our experiments, the experimental group uses the model with driving task semantic information guidance (SalM$^2$), while the control group uses the model without semantic information (Backbone network). We qualitatively compare the attention distribution predicted by different models in the same driving scenarios to determine the significant impact of driving tasks on driver attention distribution.

Due to the minimal differences between consecutive frames, we applied interval sampling and selected videos with rich scene information for analysis, extracting 6 frames as a sequence (with the first two and last two frames depicting normal scenes where the driver's attention is focused on the center of the road). Below, we analyze two scenarios.

Scenario 1: A car crosses the road. As shown in Figure \ref{fig:example1}, when a car suddenly enters the driver’s field of view from the left, our model promptly captures this semantic information and allocates attention to the car. When the car moves to the center of the field of view, the model can still make accurate predictions even without semantic information, as the driver's attention is typically concentrated in the center under most circumstances.
\begin{figure} 
    \centering
    \includegraphics[width=1\linewidth]{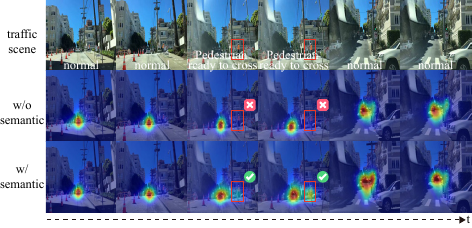}
    \caption{Visualize the driver's attention under the semantic information of a pedestrian ready to cross the road.}
    \label{fig:example2}
\end{figure}
\appendix

Scenario 2: A pedestrian is ready to cross the road. As shown in Figure \ref{fig:example2}, a pedestrian is positioned ahead on the right, potentially about to cross the road, requiring the driver to constantly monitor the pedestrian's movements. With semantic information, the model can promptly allocate attention to the pedestrian, thereby enhancing the ability to avoid traffic accidents.

Overall, our model can better understand the information of the scene and adjust the driver's attention distribution in time to improve driving safety. Meanwhile, in the scenario where the information of the scene is single and the driver's attention is distributed at the end of the road, our model can achieve the same or even better results than other models.

\begin{table} 
\centering
\fontsize{30}{40}\selectfont%设置字体 {大小}{间距}
\resizebox{\linewidth}{!}{%
\begin{tabular}{c|c|cccccc|c}
\hline
 Dataset&
  Size &
  \begin{tabular}[c]{@{}c@{}}AUC\_\\ Borji↑\end{tabular} &
  \begin{tabular}[c]{@{}c@{}}AUC\_\\ Judd↑\end{tabular} &
  NSS↑ &
  CC↑ &
  SIM↑ &
  KLD↓ &
  FLOPs(G) \\ \hline\hline
\multirow{2}{*}{TrafficGaze} & 256$^2$ & 0.92 & 0.98 & 5.90  & 0.94 & 0.78 & 0.28 & 4.45 \\
                             & 512$^2$ & 0.92 & 0.98 & 6.04 & 0.95 & 0.80  & 0.26 & 4.72 \\ \hline
\multirow{2}{*}{\begin{tabular}[c]{@{}c@{}}DrFixD\\ -rainy\end{tabular}} &
  256$^2$ &
  0.89 &
  0.95 &
  4.31 &
  0.86 &
  0.68 &
  0.47 &
  4.45 \\
                             & 512$^2$ & 0.90  & 0.96 & 4.26 & 0.86 & 0.69 & 0.45 & 4.72 \\ \hline
\multirow{2}{*}{BDDA}       & 256$^2$ & -    & -    & -    & 0.64 & 0.47 & 1.08 & 4.45 \\
                             & 512$^2$ & -    & -    & -    & 0.64 & 0.47 & 1.09 & 4.72 \\ \hline
\end{tabular}
}
\caption{Ablation study of the input image resolution. ``Size" represents the size of the input image. The symbol ``-" indicates that this metric could not be computed due to reasons related to the dataset.}
\label{tab:study_3}
\end{table}

\subsubsection{Reasonable Input Image Resolution.} Why do we use images with a smaller resolution as input? The “Top-down” branch of our network uses the CLIP model to extract semantic features. However, CLIP is limited to 224$\times$224 input image size. Even when using higher resolution images, CLIP will compress them down to this specified size, and the semantic information does not consequently increase. As shown in the table \ref{tab:study_3}, the experimental results demonstrate that: these metrics result in only a slight improvement over 256$\times$256 input images. However, it has a FLOPs of 4.72G, which significantly increases the computational amount.

Specifically, for TrafficGze and DrFixD-rainy, where the semantic information of the driving scenes is less rich, the model's performance is more dependent on the branch of ``Bottom-up". Therefore, a larger resolution will enhance the performance of the branch of ``Bottom-up", thereby improving the overall performance. For BDDA, the driving scene semantic information is very rich, and the model's performance is more dependent on branch of ``Top-down". In this case, a larger resolution will not improve performance. Instead, it will cause a significant disparity in input size between ``Bottom-up" and ``Top-down", leading to a mismatch between semantic information and image features. Consequently, a decline in model performance, and we ultimately decide to choose a resolution of 256$\times$256 as the input.

\section{Conclusion}
In driving, a driver's attention is predominantly focused on the area ahead, such as the road's endpoint or the vehicle in front. However, in complex environments, this attention can shift due to sudden events in the field of view, making the semantic information of the current driving scene crucial for understanding driver attention. We have investigated driver attention through the lens of dual-process visual cognition and developed a driver attention prediction model based on driving scene information. This model integrates ``Top-down" semantic information (high-level) with ``Bottom-up" image information (low-level) for effective attention prediction.

Extensive experiments demonstrate that guiding model understanding of driver attention through driving scene information can be an effective approach. Our model achieves SOTA or near-SOTA performance with minimal parameters. In future work, we aim to explore more effective methods for using driving scene semantic information to guide driver attention prediction, further improving prediction accuracy.

\section{Acknowledgments}
This work was supported by the National Natural Science Foundation of China (62106208, 62373311), Natural Science Foundation of Sichuan Province under Grant 2025ZNSFSC0471, 2023NSFSC1418, and in part by the China Postdoctoral Science Foundation under Grant 2021TQ0272, 2021M702715.

\bibliography{aaai25}

\begin{thebibliography}{38}
\providecommand{\natexlab}[1]{#1}

\bibitem[{Alkaabi(2023)}]{alkaabi2023identification}
Alkaabi, K. 2023.
\newblock Identification of hotspot areas for traffic accidents and analyzing drivers’ behaviors and road accidents.
\newblock \emph{Transportation research interdisciplinary perspectives}, 22: 100929.

\bibitem[{Alletto et~al.(2016)Alletto, Palazzi, Solera, Calderara, and Cucchiara}]{alletto2016dr}
Alletto, S.; Palazzi, A.; Solera, F.; Calderara, S.; and Cucchiara, R. 2016.
\newblock Dr (eye) ve: a dataset for attention-based tasks with applications to autonomous and assisted driving.
\newblock In \emph{Proceedings of the ieee conference on computer vision and pattern recognition workshops}, 54--60.

\bibitem[{Baee et~al.(2021)Baee, Pakdamanian, Kim, Feng, Ordonez, and Barnes}]{baee2021medirl}
Baee, S.; Pakdamanian, E.; Kim, I.; Feng, L.; Ordonez, V.; and Barnes, L. 2021.
\newblock Medirl: Predicting the visual attention of drivers via maximum entropy deep inverse reinforcement learning.
\newblock In \emph{Proceedings of the IEEE/CVF international conference on computer vision}, 13178--13188.

\bibitem[{Balasubramani et~al.(2024)Balasubramani, Aravindhar, Renjith, and Ramesh}]{balasubramani2024ddss}
Balasubramani, S.; Aravindhar, J.; Renjith, P.; and Ramesh, K. 2024.
\newblock DDSS: Driver decision support system based on the driver behaviour prediction to avoid accidents in intelligent transport system.
\newblock \emph{International Journal of Cognitive Computing in Engineering}, 5: 1--13.

\bibitem[{Borji, Sihite, and Itti(2012)}]{borji2012quantitative}
Borji, A.; Sihite, D.~N.; and Itti, L. 2012.
\newblock Quantitative analysis of human-model agreement in visual saliency modeling: A comparative study.
\newblock \emph{IEEE Transactions on Image Processing}, 22(1): 55--69.

\bibitem[{Brishtel et~al.(2022)Brishtel, Krauß, Schmidt, Rambach, Vozniak, and Stricker}]{9945234}
Brishtel, I.; Krauß, S.; Schmidt, T.; Rambach, J.~R.; Vozniak, I.; and Stricker, D. 2022.
\newblock Classification of Manual Versus Autonomous Driving based on Machine Learning of Eye Movement Patterns.
\newblock In \emph{2022 IEEE International Conference on Systems, Man, and Cybernetics (SMC)}, 700--705.

\bibitem[{Chen, Nan, and Xiang(2023)}]{chen2023fblnet}
Chen, Y.; Nan, Z.; and Xiang, T. 2023.
\newblock FBLNet: FeedBack Loop Network for Driver Attention Prediction.
\newblock In \emph{Proceedings of the IEEE/CVF International Conference on Computer Vision}, 13371--13380.

\bibitem[{Cornia et~al.(2016)Cornia, Baraldi, Serra, and Cucchiara}]{cornia2016deep}
Cornia, M.; Baraldi, L.; Serra, G.; and Cucchiara, R. 2016.
\newblock A deep multi-level network for saliency prediction.
\newblock In \emph{2016 23rd International Conference on Pattern Recognition (ICPR)}, 3488--3493. IEEE.

\bibitem[{Deng et~al.(2024)Deng, Jiang, Shi, Wu, Wu, Yan, Zhang, and Yan}]{deng2023driving}
Deng, T.; Jiang, L.; Shi, Y.; Wu, J.; Wu, Z.; Yan, S.; Zhang, X.; and Yan, H. 2024.
\newblock Driving Visual Saliency Prediction of Dynamic Night Scenes via a Spatio-Temporal Dual-Encoder Network.
\newblock \emph{IEEE Transactions on Intelligent Transportation Systems}, 25(3): 2413--2423.

\bibitem[{Deng, Yan, and Yan(2021)}]{9428151}
Deng, T.; Yan, F.; and Yan, H. 2021.
\newblock Driving Video Fixation Prediction Model Via Spatio-Temporal Networks and Attention Gates.
\newblock In \emph{2021 IEEE International Conference on Multimedia and Expo (ICME)}, 1--6.

\bibitem[{Deng, Yan, and Li(2018)}]{deng2017learning}
Deng, T.; Yan, H.; and Li, Y.-J. 2018.
\newblock Learning to boost bottom-up fixation prediction in driving environments via random forest.
\newblock \emph{IEEE Transactions on Intelligent Transportation Systems}, 19(9): 3059--3067.

\bibitem[{Deng et~al.(2019)Deng, Yan, Qin, Ngo, and Manjunath}]{deng2019drivers}
Deng, T.; Yan, H.; Qin, L.; Ngo, T.; and Manjunath, B. 2019.
\newblock How do drivers allocate their potential attention? Driving fixation prediction via convolutional neural networks.
\newblock \emph{IEEE Transactions on Intelligent Transportation Systems}, 21(5): 2146--2154.

\bibitem[{Du, Deng, and Yan(2023)}]{du2023causes}
Du, P.; Deng, T.; and Yan, F. 2023.
\newblock What Causes a Driver's Attention Shift? A Driver's Attention-Guided Driving Event Recognition Model.
\newblock In \emph{2023 International Joint Conference on Neural Networks (IJCNN)}, 1--8. IEEE.

\bibitem[{Evans et~al.(2011)Evans, Horowitz, Howe, Pedersini, Reijnen, Pinto, Kuzmova, and Wolfe}]{evans2011visual}
Evans, K.~K.; Horowitz, T.~S.; Howe, P.; Pedersini, R.; Reijnen, E.; Pinto, Y.; Kuzmova, Y.; and Wolfe, J.~M. 2011.
\newblock Visual attention.
\newblock \emph{Wiley Interdisciplinary Reviews: Cognitive Science}, 2(5): 503--514.

\bibitem[{Fang et~al.(2024)Fang, Wang, Xue, and Chua}]{fang2024behavioral}
Fang, J.; Wang, F.; Xue, J.; and Chua, T.-S. 2024.
\newblock Behavioral Intention Prediction in Driving Scenes: A Survey.
\newblock \emph{IEEE Transactions on Intelligent Transportation Systems}, 25(8): 8334--8355.

\bibitem[{Fang et~al.(2019)Fang, Yan, Qiao, and Xue}]{fang2019dada}
Fang, J.; Yan, D.; Qiao, J.; and Xue, J. 2019.
\newblock Dada: A large-scale benchmark and model for driver attention prediction in accidental scenarios.
\newblock \emph{arXiv preprint arXiv:1912.12148}, 3.

\bibitem[{Fang et~al.(2021)Fang, Yan, Qiao, Xue, and Yu}]{fang2021dada}
Fang, J.; Yan, D.; Qiao, J.; Xue, J.; and Yu, H. 2021.
\newblock DADA: Driver attention prediction in driving accident scenarios.
\newblock \emph{IEEE transactions on intelligent transportation systems}, 23(6): 4959--4971.

\bibitem[{Fang et~al.(2022)Fang, Yan, Qiao, Xue, and Yu}]{fang2022dada}
Fang, J.; Yan, D.; Qiao, J.; Xue, J.; and Yu, H. 2022.
\newblock DADA: Driver Attention Prediction in Driving Accident Scenarios.
\newblock \emph{IEEE Transactions on Intelligent Transportation Systems}, 23(6): 4959--4971.

\bibitem[{Fu et~al.(2019)Fu, Liu, Tian, Li, Bao, Fang, and Lu}]{fu2019dual}
Fu, J.; Liu, J.; Tian, H.; Li, Y.; Bao, Y.; Fang, Z.; and Lu, H. 2019.
\newblock Dual attention network for scene segmentation.
\newblock In \emph{Proceedings of the IEEE/CVF conference on computer vision and pattern recognition}, 3146--3154.

\bibitem[{Gu and Dao(2023)}]{gu2023mamba}
Gu, A.; and Dao, T. 2023.
\newblock Mamba: Linear-time sequence modeling with selective state spaces.
\newblock \emph{arXiv preprint arXiv:2312.00752}.

\bibitem[{Hertz-Palmor et~al.(2023)Hertz-Palmor, Rozenblit, Lavi, Zeltser, Kviatek, and Lazarov}]{hertz2023aberrant}
Hertz-Palmor, N.; Rozenblit, D.; Lavi, S.; Zeltser, J.; Kviatek, Y.; and Lazarov, A. 2023.
\newblock Aberrant reward learning, but not negative reinforcement learning, is related to depressive symptoms: an attentional perspective.
\newblock \emph{Psychological Medicine}, 1--14.

\bibitem[{Itti and Koch(2001)}]{itti2001computational}
Itti, L.; and Koch, C. 2001.
\newblock Computational modelling of visual attention.
\newblock \emph{Nature reviews neuroscience}, 2(3): 194--203.

\bibitem[{Judd et~al.(2009)Judd, Ehinger, Durand, and Torralba}]{judd2009learning}
Judd, T.; Ehinger, K.; Durand, F.; and Torralba, A. 2009.
\newblock Learning to predict where humans look.
\newblock In \emph{2009 IEEE 12th international conference on computer vision}, 2106--2113. IEEE.

\bibitem[{Kotseruba and Tsotsos(2024)}]{kotseruba2024scout+}
Kotseruba, I.; and Tsotsos, J.~K. 2024.
\newblock SCOUT+: Towards Practical Task-Driven Drivers' Gaze Prediction.
\newblock \emph{arXiv preprint arXiv:2404.08756}.

\bibitem[{Le~Meur, Le~Callet, and Barba(2007)}]{le2007predicting}
Le~Meur, O.; Le~Callet, P.; and Barba, D. 2007.
\newblock Predicting visual fixations on video based on low-level visual features.
\newblock \emph{Vision research}, 47(19): 2483--2498.

\bibitem[{Min and Corso(2019)}]{min2019tased}
Min, K.; and Corso, J.~J. 2019.
\newblock Tased-net: Temporally-aggregating spatial encoder-decoder network for video saliency detection.
\newblock In \emph{Proceedings of the IEEE/CVF International Conference on Computer Vision}, 2394--2403.

\bibitem[{Peters et~al.(2005)Peters, Iyer, Itti, and Koch}]{peters2005components}
Peters, R.~J.; Iyer, A.; Itti, L.; and Koch, C. 2005.
\newblock Components of bottom-up gaze allocation in natural images.
\newblock \emph{Vision research}, 45(18): 2397--2416.

\bibitem[{Qin et~al.(2022)Qin, Shi, He, Zhang, Zhang, Li, Deng, and Yan}]{qin2022id}
Qin, L.; Shi, Y.; He, Y.; Zhang, J.; Zhang, X.; Li, Y.; Deng, T.; and Yan, H. 2022.
\newblock ID-YOLO: Real-time salient object detection based on the driver’s fixation region.
\newblock \emph{IEEE Transactions on Intelligent Transportation Systems}, 23(9): 15898--15908.

\bibitem[{Radford et~al.(2021)Radford, Kim, Hallacy, Ramesh, Goh, Agarwal, Sastry, Askell, Mishkin, Clark, Krueger, and Sutskever}]{radford2021learningtransferablevisualmodels}
Radford, A.; Kim, J.~W.; Hallacy, C.; Ramesh, A.; Goh, G.; Agarwal, S.; Sastry, G.; Askell, A.; Mishkin, P.; Clark, J.; Krueger, G.; and Sutskever, I. 2021.
\newblock Learning Transferable Visual Models From Natural Language Supervision.
\newblock arXiv:2103.00020.

\bibitem[{Rasouli and Tsotsos(2018)}]{rasouli2018joint}
Rasouli, A.; and Tsotsos, J.~K. 2018.
\newblock Joint attention in driver-pedestrian interaction: from theory to practice.
\newblock \emph{arXiv preprint arXiv:1802.02522}.

\bibitem[{Riche et~al.(2013)Riche, Duvinage, Mancas, Gosselin, and Dutoit}]{riche2013saliency}
Riche, N.; Duvinage, M.; Mancas, M.; Gosselin, B.; and Dutoit, T. 2013.
\newblock Saliency and human fixations: State-of-the-art and study of comparison metrics.
\newblock In \emph{Proceedings of the IEEE international conference on computer vision}, 1153--1160.

\bibitem[{Shen et~al.(2022)Shen, Wijayaratne, Sriram, Hasan, Du, and Driggs-Campbell}]{shen2022cocatt}
Shen, Y.; Wijayaratne, N.; Sriram, P.; Hasan, A.; Du, P.; and Driggs-Campbell, K. 2022.
\newblock CoCAtt: A cognitive-conditioned driver attention dataset.
\newblock In \emph{2022 IEEE 25th International Conference on Intelligent Transportation Systems (ITSC)}, 32--39. IEEE.

\bibitem[{Shi et~al.(2024)Shi, Zhao, Wu, Wu, and Yan}]{shi2023fixated}
Shi, Y.; Zhao, S.; Wu, J.; Wu, Z.; and Yan, H. 2024.
\newblock Fixated Object Detection Based on Saliency Prior in Traffic Scenes.
\newblock \emph{IEEE Transactions on Circuits and Systems for Video Technology}, 34(3): 1413--1426.

\bibitem[{Tawari et~al.(2014)Tawari, M{\o}gelmose, Martin, Moeslund, and Trivedi}]{tawari2014attention}
Tawari, A.; M{\o}gelmose, A.; Martin, S.; Moeslund, T.~B.; and Trivedi, M.~M. 2014.
\newblock Attention estimation by simultaneous analysis of viewer and view.
\newblock In \emph{17th International IEEE Conference on Intelligent Transportation Systems (ITSC)}, 1381--1387. IEEE.

\bibitem[{Tian, Deng, and Yan(2022)}]{tian2022driving}
Tian, H.; Deng, T.; and Yan, H. 2022.
\newblock Driving as well as on a sunny day? predicting driver's fixation in rainy weather conditions via a dual-branch visual model.
\newblock \emph{IEEE/CAA Journal of Automatica Sinica}, 9(7): 1335--1338.

\bibitem[{Vozniak et~al.(2023)Vozniak, M{\"u}ller, Hell, Lipp, Abouelazm, and M{\"u}ller}]{vozniak2023context}
Vozniak, I.; M{\"u}ller, P.; Hell, L.; Lipp, N.; Abouelazm, A.; and M{\"u}ller, C. 2023.
\newblock Context-empowered visual attention prediction in pedestrian scenarios.
\newblock In \emph{Proceedings of the IEEE/CVF Winter Conference on Applications of Computer Vision}, 950--960.

\bibitem[{Wu et~al.(2024)Wu, Liu, Liang, and Chang}]{wu2024ultralight}
Wu, R.; Liu, Y.; Liang, P.; and Chang, Q. 2024.
\newblock Ultralight vm-unet: Parallel vision mamba significantly reduces parameters for skin lesion segmentation.
\newblock \emph{arXiv preprint arXiv:2403.20035}.

\bibitem[{Xia et~al.(2019)Xia, Zhang, Kim, Nakayama, Zipser, and Whitney}]{xia2019predicting}
Xia, Y.; Zhang, D.; Kim, J.; Nakayama, K.; Zipser, K.; and Whitney, D. 2019.
\newblock Predicting driver attention in critical situations.
\newblock In \emph{Computer Vision--ACCV 2018: 14th Asian Conference on Computer Vision, Perth, Australia, December 2--6, 2018, Revised Selected Papers, Part V 14}, 658--674. Springer.

\end{thebibliography}

\end{document}